# HIGHLIGHTING OBJECTS OF INTEREST IN AN IMAGE BY INTEGRATING SALIENCY AND DEPTH

*Subhayan Mukherjee, Irene Cheng and Anup Basu*

Department of Computing Science, University of Alberta, Edmonton, Canada
Emails: {mukherje,locheng,basu} ualberta.ca

## ABSTRACT

Stereo images have been captured primarily for 3D reconstruction in the past. However, the depth information acquired from stereo can also be used along with saliency to highlight certain objects in a scene. This approach can be used to make still images more interesting to look at, and highlight objects of interest in the scene. We introduce this novel direction in this paper, and discuss the theoretical framework behind the approach. Even though we use depth from stereo in this work, our approach is applicable to depth data acquired from any sensor modality. Experimental results on both indoor and outdoor scenes demonstrate the benefits of our algorithm.

## 1. INTRODUCTION

Depth-of-Field (DoF) is an essential component in producing photorealistic effects during image capture. DoF enhances images not only by giving people the "feeling" of depth information but also allows them to focus on the important regions. DoF makes images look more natural. It is essential in making the focus points of the image stand out, by emphasizing the foreground and de-emphasizing the background [1].

While taking a photo with an optical camera, we can vary the size of the aperture to set the DoF or "zone of focus" for the photo. The points within the DoF appear to be focused, while other points far away from the focal plane are blurred. Photos with a small zone of focus are said to have a "shallow" DoF. Thus, to highlight objects of interest in a photo, we should select a shallow DoF containing only these objects, also known as "Region-of-Interest" (ROI) or "foreground." However, there is often a requirement to render shallow DoF effects during post-processing, such as during photo retouching by a professional photographer in his or her studio, or during cinematographic editing sessions. Furthermore, in the areas of virtual reality and video games, real-time DoF effects are also important aspects in enhancing the visual effects. Thus, simulating DoF effects have become an important research topic in the field of computer vision.

Most DoF rendering algorithms post-process single-view images, and can be roughly categorized as object-space-based and image-space-based methods. Lee's method [2] performs image blurring via non-linear interpolation of mipmap images generated from a pinhole image. Multilayer approaches like our proposed method split an image into layers based on pixel depths. In others, hidden image areas are approximated by color extrapolation to solve the partial visibility problem via Fourier transform, pyramid image processing, anisotropic diffusion, splatting, rectangle spreading, and so on [1].

Saliency is widely used to investigate visual attention. The biologically inspired method by Itti [3] determines image saliency using Difference-of-Gaussians. Itti's method was later extended with graph-based normalization to build the visual saliency model [4]. Other methods use frequency domain processing, or combine global contrast and spatial relationship of pixels to detect entire salient objects [1].

Depth estimation from stereo images is a well-researched problem [5]. Following decades of research on computational stereo, modern-day stereo algorithms are capable of producing reasonably accurate depth estimates, sometimes with real-time performance [6]. So, we mostly use stereo depth maps in our proposed work, although our proposed method will work with depth maps acquired from other modalities as well. Furthermore, many elements of stereo algorithms are now well understood; particularly, accurate stereo calibration and efficient algorithms for local correspondence. Consequently, research focus has now shifted to the more difficult problems of stereopsis like global correspondence and occlusion handling.

Stereo depth estimation approaches can be broadly classified as local and global methods [5], according the type of constraints they exploit while attempting to match pixels in one image with their corresponding pixels in the other image. Constraints on a small number of pixels surrounding a pixel of interest are referred to as "local" constraints, and constraints on scan-lines or on the entire image are termed as "global" constraints. Thus, local methods use block matching, gradient-based optimization or feature matching to find correspondences, whereas global methods use dynamic programming, intrinsic curves, graph cuts, non-linear diffusion, belief propagation, etc. Local methods which perform block-matching employ several well-known "match metrics" for computing the similarity between blocks in the left and right images. These metrics include normalized cross-correlation

(NCC), sum of squared differences (SSD), normalized SSD, sum of absolute differences (SAD), rank, and census.

Recent stereo methods capable of handling occlusions are classified into three broad categories as follows. Some methods simply detect occlusions using depth map discontinuities, left-right matching or intensity edges. Other methods reduce sensitivity of the stereo matching process to occlusions, using robust similarity criterion or adaptive regions of support. Lastly, there are methods that model occlusion geometry via global occlusion modelling, multiple cameras or active vision.

The rest of this paper is organized as follows: Section 2 motivates the proposed architecture. Section 3 explains the theoretical model used along with its implementation. Section 4 presents a visual comparison of the performance of the proposed architecture against existing methods which rely on manual ROI selection by the user. Finally, Section 5 presents some concluding remarks and directions for future work.

## 2. MOTIVATION AND PROPOSED ARCHITECTURE

Review of literature reveals that even the most recent depth of field rendering techniques require manual interaction for selecting the depth ranges of interest to the user [7, 6]. We aim to eliminate this manual step, thereby making the process fully automatic. One such automated approach [8] exists, which uses saliency itself to compute the depth map. There is also an enhanced extended version [1] of that algorithm which uses a pair of images with and without flash to get more accurate depth extraction and DoF region extraction. However, their results show successful DoF simulation only for macro subjects, instead of the more "general" case where the ROI is at an arbitrary distance from the camera. They also show that in images like Fig.3(j), where non-ROI (background) pixels have varying (true) depth, albeit similar saliency values, their depth estimation fails. This causes poor DoF simulation, as they end up blurring all background pixels equally.

It should be noted that our proposed framework does not impose any restriction on how the depth map or the saliency map should be obtained. Thus, future research could investigate using different types of depth and saliency extraction techniques in conjunction with our proposed framework.

More specifically, this paper proposes novel methods to:

1. Automatically determine the user's "depths of interest" based on the range of depths lying in the regions of interest (ROIs) of the image, which, in turn is determined from the supplied depth and saliency information.

2. Compute a "defocus" map which assigns a defocus, or blur level to each image pixel, which depends on:

    (a) The pixel's distance (in terms of depth) from the range of depths corresponding to the ROI.

    (b) The probability distribution of depth levels among image pixels, such that the DoF effect is clearly visible even when the range of depths of interest is comparable to the range of possible depth levels.

The salient regions of the image are first determined by thresholding the saliency map [9]. Then, by using our proposed novel methods, we first filter the salient regions based on the depth map, thereby rejecting those regions which presumably lie at depths beyond the user's ROIs. Then, we build the defocus map by mapping each pixel to a defocus (blur) level. Fig.1 shows an overview of our proposed framework.

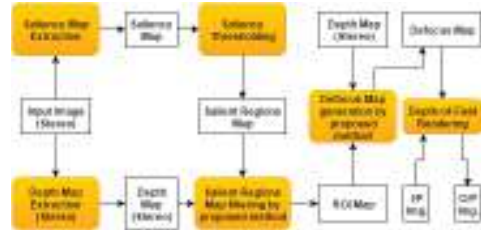

**Fig. 1**. An overview of our proposed framework.

## 3. THEORETICAL MODEL AND IMPLEMENTATION

Fig.1 clarifies the above discussion; i.e., that the core functionality of our proposed method lies in filtering the salient regions of the image based on their depths, and subsequently using the output of this step along with the image's depth map to determine the defocus (blur) level of each image pixel.

We assume that the user's objects (or, regions) of interest lie at close proximity to each other with respect to their depths (Z values). We call *this* range of depth "depths of interest." However, we do not assume that the ROIs are nearby each other w.r.t their X or Y axes. Also, we do not assume that the ROIs lie very close to the camera, like in macro photography.

Let $D_s$ be the ordered set of depths of pixels comprising the salient regions, as detected by saliency thresholding [9]. Then, $\Delta D_s$, the forward difference of $D_s$, gives us the distance of separation between consecutive depth levels in $D_s$.

Now as per our above assumption, we need to determine a "proximity threshold," $T_p$ for the depths in $D_s$. This is done by normalizing $\Delta D_s$ and running Ostu's method [10] on it. Then, the *first* set of consecutive depth values in $D_s$ which are separated by a distance more than $T_p$ mark the *boundary* of the range of depths of interest, $F$.

Next, we assign a defocus level to each pixel. The central idea is that pixels lying within the depth range $F$ should be in focus. All other pixels will have a defocus amount proportional to their distance (in terms of depth) from the nearest depth level in $F$. This translates to an output image with a sharp focus on the depths of user's interest and gradually increasing defocus on depths farther away. However, this implies that for images which have a low range of possible depth levels, defocus of non-interest regions will be hardly visible if the range of depths of interest is substantially large, or comparable to the range of possible depth levels in that image.

A possible solution to the aforementioned problem is to adjust the blur level of all pixels to be defocused, by using an image-dependent constant factor, in a way that in images with low depth range the blur level is increased by a greater degree compared to those in which the depth range is relatively high. However, this means that even if very few pixels have abnormally high (or low) depth values compared to all others, the blur level adjustment of all other pixels will suffer. Thus, a "reasonable" blur level adjustment policy must take into account the *probability distribution* of all pixels' depths.

To mathematically express the blur level computation for image pixels (defocus map) by our proposed method based on the above discussions, we first introduce the following terms (please note that the superscript $L$ simply denotes "linear"):

$$b_p^L = \min_{\forall f \in F} |f - d_p| \quad (1)$$

So, $b_p^L$ denotes the (linearly increasing) blur level for a pixel $p$ belonging to the input image $I$ and having a depth value $d_p$, and $f$ denotes the individual depth levels constituting $F$.

$$B^L = \{(p, b_p^L) \mid p \in I\} \quad (2)$$

$B^L$ maps all input image pixels to their linear blur levels.

$$T = \{t := b_p^L \mid (p, b_p^L) \in B^L \wedge b_p^L > 0\} \quad (3)$$

$T$ is the set of (distinct) positive linear blur levels in $I$.

$$C_t = \{p \mid (p, b_p^L) \in B^L \wedge t := b_p^L \in T\} \quad (4)$$

$C_t$ denotes the set of pixels in $I$ having linear blur level $t$.

$$C_T = \bigcup_{t \in T} C_t \quad (5)$$

$C_T$ is the set of pixels in $I$ having positive linear blur levels.

$$\alpha_I = \frac{1}{255} * \sum_{t \in T} \frac{|C_t|}{|C_T|} * t \quad (6)$$

$\alpha_I$ is the normalized blur adjustment factor for $I$, based on the probability distribution of $b_p^L$ values for all pixels in $C_T$. It can be shown that all possible values for $\alpha_I$ lie within (0, 255), so we normalize $\alpha_I$ to (0, 1) by dividing it by 255. Thus, $\alpha_I$ is essentially the *mathematical expectation* of positive $b_p^L$ values for I, and we normalize this expected value to (0, 1).

$$\sigma_t = \begin{cases} \alpha_I * t & \text{if } \alpha_I < \beta \\ \frac{1}{\gamma} * \alpha_I * t & \text{otherwise; } \gamma > 1.0 \end{cases} \quad (7)$$

$\sigma_t$ is the value of the standard deviation for the Gaussian kernel used to blur any pixel of $I$ with a linear blur level $t$. $\beta$, $\gamma$ are additional parameters that control the strength of the blur (defocus). The "defocus map" maps each pixel to its $\sigma_t$ value.

Essentially, Eq. 7 enforces the constraint that, if the blur-adjustment factor for image $I$, $\alpha_I$ is lesser than the threshold $\beta$, (i.e. range of Depths of Interest is comparable to the range of all possible depths in $I$) then we defocus the non-interest regions strongly, otherwise we *damp* it by a factor $\gamma > 1.0$.

## 4. EXPERIMENTAL RESULTS

We first present a visual comparison of the performance of our algorithm against two recent papers which rely on manual marking of the ROI by the user. We use the fixed values $\beta = 0.2$, $\gamma = 5.0$ for all images, as they produced the best results.

The depth maps were obtained from the same sources as their corresponding input images, and the saliency maps were generated using a Graph-based Visual Saliency algorithm [4].

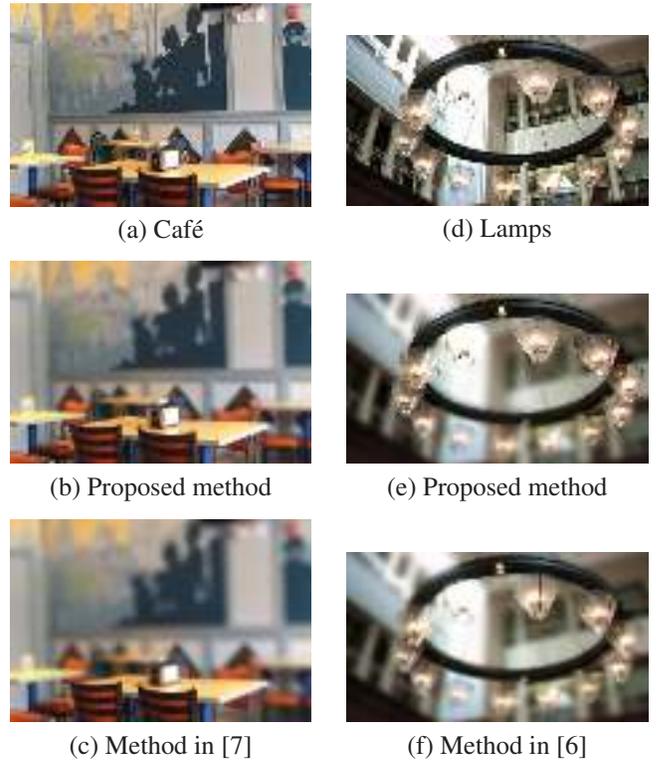

(a) Café  (d) Lamps

(b) Proposed method  (e) Proposed method

(c) Method in [7]  (f) Method in [6]

**Fig. 2**. Performance comparison on indoor scenes. The box in Café and the nearest Lamps are in better focus in our output.

Comparisons shown in Fig.2 and Fig.3 illustrate that our algorithm produces results as good as those which rely on manual marking of ROI by the user [7, 6] even without varying any of its parameter values. Moreover, our method is fully automatic and does not rely on any user intervention. Also, in Table 1, we have indicated the computed value of $\alpha_I$ for each output of our algorithm. This justifies the motivation behind choosing the value of $\beta = 0.2$: by doing this, scenes having a *low* expected $b_p^L$ value ($\alpha_I < 0.2$) get a *steeper* increase in blur with increasing distance (depth) from the ROI, while for others, we "damp" the *slope* of this increment by a factor $\gamma$.

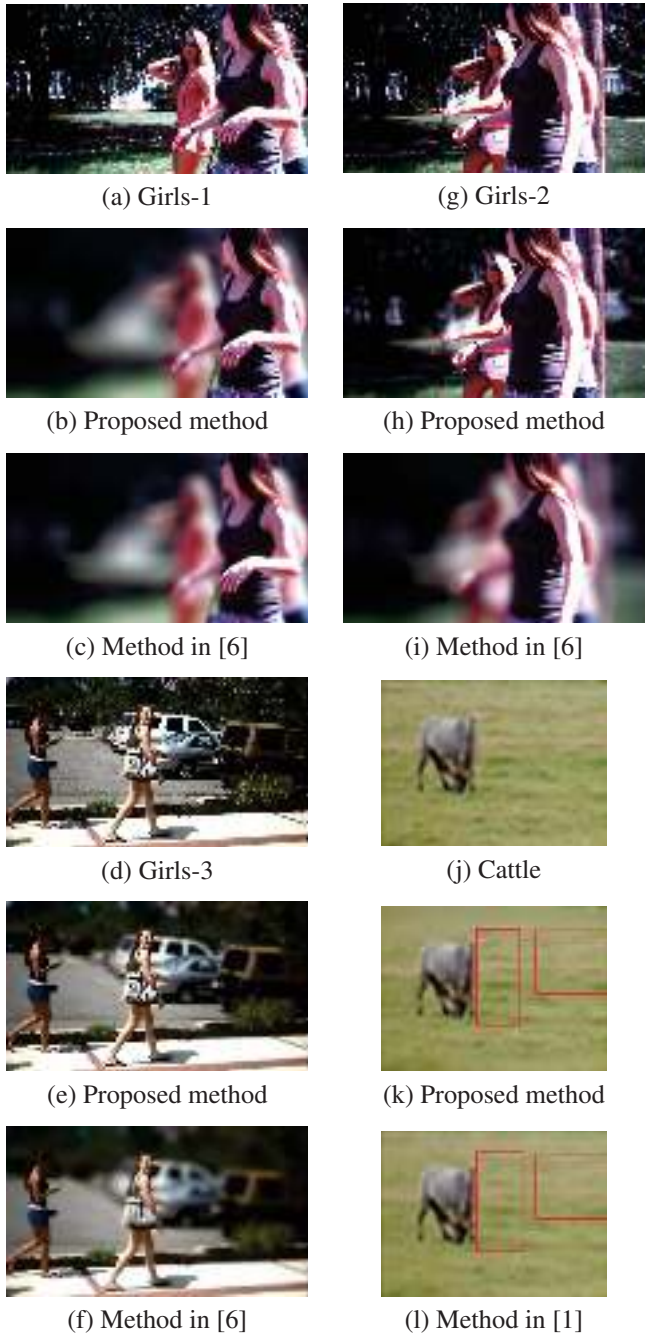

| Input image | No. of pixels | Time (s) | Computed $\alpha_I$ |
|---|---|---|---|
| Cattle | 22272 | 0.145 | 0.13 |
| Cafe | 642866 | 2.235 | 0.15 |
| Girls-1 | 230400 | 0.601 | 0.53 |
| Girls-2 | 230400 | 0.701 | 0.24 |
| Girls-3 | 230400 | 0.533 | 0.22 |
| Lamps | 230400 | 0.527 | 0.16 |
| Average | 264456 | 0.780 | |

**Table 1**. Execution time and $\alpha_I$'s computed by our method.

(a) Girls-1
(g) Girls-2
(b) Proposed method
(h) Proposed method
(c) Method in [6]
(i) Method in [6]
(d) Girls-3
(j) Cattle
(e) Proposed method
(k) Proposed method
(f) Method in [6]
(l) Method in [1]

**Fig. 3**. Performance comparison on outdoor scenes. The right hand of the girl in front is in better focus in our output in Girls-1 and Girls-2. In Girls-3, we can see the smoother increase in blur in our output for scenes having a wider variance in depth. Cattle shows that our method performs well even when most background pixels have similar saliency values. So, we do not suffer from problems faced in similar scenarios by [8, 1]. Image areas marked in red squares show that our method applies lesser defocus to background pixels lying closer (depth-wise) to the ROI, but [8, 1] blurs all background pixels equally.

The image "Cattle" shows that our method performs well even when most background pixels have similar saliency values, and thus we do not suffer from problems faced in similar scenarios by [8, 1]. Monocular depth map for the image Cattle was extracted[1] using methods described in [11, 12, 13].

It should also be noted that we do not compare our method specifically with [8], but only with [1], because the latter is an extension of the former and claims to produce better results.

Table 1 shows the time taken by proposed method to filter the map of salient regions and generate the defocus map. We used a PC running Matlab R2015b on Windows 7 on a 2.26 GHz Intel core $i3$ processor with 4 GB RAM. No data- or instruction- parallelism was used in the implementation. By comparing this with the methods [8, 1] it can be observed that they take over 2 seconds to extract and refine the DoF region even for a smaller (90000 pixels) image on a faster PC.

## 5. CONCLUSION AND FUTURE WORK

A fully automated framework of rendering shallow Depth-of-Field effects in post-processing, using depth and saliency information was proposed. Our method automatically detects the ROI for the input image based on the computed depth and saliency information. Experiments show that even with fixed parameter values our framework produces output of similar quality as existing methods which rely on manual marking of the ROI by the user. Our method does not suffer from a limitation faced by another automated method, namely poor DoF rendering for images where non-ROI (background) pixels have varying (true) depth, but similar saliency [8, 1].

In future work we will focus on automatic estimation of the value of parameter $\beta$ used in our algorithm. We will also investigate applications in foveation [14, 15, 16, 17, 18, 19], model based coding [20, 21, 22], and medical imaging [23, 24, 25].

## 6. ACKNOWLEDGEMENTS

The authors thank Alberta Innovates and NSERC, Canada, for their financial support of this research.

---
[1]http://make3d.cs.cornell.edu